\documentclass[10pt,twocolumn,letterpaper]{article}

\usepackage[pagenumbers]{wacv}       %

\usepackage{booktabs}
\usepackage{multirow}
\usepackage{graphicx}
\graphicspath{{fig/}}
\newcommand{\probeRange}{0.75--0.91}
\newcommand{\zeroshotRange}{0.48--0.66}

\definecolor{wacvblue}{rgb}{0.21,0.49,0.74}
\usepackage[pagebackref,breaklinks,colorlinks,allcolors=wacvblue]{hyperref}

\def\wacvPaperID{*****}
\def\confName{WACV}
\def\confYear{2027}

\title{Whose Face Is It Anyway? A Multi-Model Audit of Facial Affect
Recognition on Children, and Why the Gap Is the Head, Not the Features}

\author{Tobias Hallmen$^{1}$ \qquad Robin-Nico Kampa$^{2}$ \qquad Elisabeth Andr\'e$^{1}$\\[2pt]
$^{1}$Chair for Human-Centered Artificial Intelligence, University of Augsburg\\
$^{2}$Institute for Distributed Intelligent Systems, University of the Bundeswehr Munich\\[2pt]
{\tt\small tobias.hallmen@uni-a.de \quad robin-nico.kampa@unibw.de \quad elisabeth.andre@uni-a.de}
}

\begin{document}
\maketitle
\begin{abstract}
Facial affect models are trained almost entirely on adults, yet are increasingly
applied to children in education, health, and developmental research. We present a
controlled, multi-model audit of five AffectNet-pretrained expression models
(EmoNet, EmotiEffLib, DDAMFN++, OpenFace~3.0, LibreFace) on children, across
four child image datasets, the AffectNet-8 validation set, and two spontaneous
child video datasets, through one shared harness. Three findings emerge. First, the
child gap is \emph{model-agnostic}: every architecture degrades from posed to
naturalistic faces and shares the fear$\rightarrow$surprise confusion. Second, it is
\emph{concentrated and corroborated} across all five models: open-mouth faces (read as
surprise, correlating with the AU26 jaw drop) and South-Asian children degrade
systematically, with a smaller averted-gaze penalty, while closed-mouth faces, White and
Black children, and direct gaze do not; the bias tracks expression morphology and specific
populations, not skin tone. Third, the gap is \emph{diagnosable}: a linear probe on frozen
features reaches \probeRange{} on unseen children versus \zeroshotRange{} zero-shot, so it
lies largely in the \emph{classifier head, not the representation}, whereas dimensional
valence/arousal regression degrades sharply under domain shift. Building on this,
recalibrating only the head on a little target data recovers $+0.13$ to $+0.28$ on the two
largest child sets across all five models at negligible adult cost, though the gain is
in-distribution and does not transfer across child collections. We will release the harness,
per-sample predictions, and analysis code; the child face data stays license-locked and is
never redistributed.
\end{abstract}

\section{Introduction}
\label{sec:intro}

Automatic facial expression recognition (FER) and dimensional affect estimation
(valence/arousal)~\cite{russell1980circumplex} are increasingly deployed on \emph{children}: in learning analytics~\cite{suemer:et:al:2023,savchenko:et:al:2022},
screening for developmental and affective conditions~\cite{trinhammer:et:al:2026}, and human--robot interaction~\cite{filippini:merla:2023}.
Yet the datasets and pretrained models that power these systems are built almost
entirely from \emph{adult} faces: AffectNet~\cite{mollahosseini2017affectnet},
RAF-DB~\cite{li2017rafdb}, and the in-the-wild affect corpora are adult-dominated, and the
backbones (VGGFace2~\cite{cao2018vggface2}, MS-Celeb-1M~\cite{guo2016msceleb},
FFHQ~\cite{karras2019ffhq}) more so.
Children are not small adults: facial morphology, expression dynamics, and emotional competencies
evolve substantially throughout development, while emotional expressions are shaped by social and
cognitive development~\cite{loechner:schuller:2022}. Consequently, children's facial expressions
differ from those represented in the adult benchmarks on which most FER models are trained. Deploying
adult-trained affect models on children therefore risks silent, demographically structured failure.

Prior child-FER evaluations are typically \emph{single-model} and \emph{single-dataset}
\cite{vivekananthan2024children,virgolin2022adults,kim2021agebias,gayamorey2025agegap},
which conflates a particular model's quirks with the underlying domain gap, and offers no
purchase on \emph{why} models fail or \emph{how} to fix them. We instead run a controlled,
\emph{multi-model} audit: five widely used, openly available AffectNet-pretrained
models (EmoNet~\cite{toisoul2021emonet}, EmotiEffLib/HSEmotion~\cite{savchenko2021hsemotion},
DDAMFN++~\cite{zhang2023ddamfn}, OpenFace~3.0~\cite{openface3}, and
LibreFace~\cite{chang2024libreface}), evaluated through a \emph{single shared harness}
(common label space, per-model native preprocessing, identical metrics) across four child
image datasets, the standard AffectNet-8 validation set, and two spontaneous child video
datasets. Because all models see the same data and harness, differences reflect the models
together with their native preprocessing (which we hold fixed per model and treat as part of each
system), and \emph{agreement} across models points to a shared cause in the data and the
(adult) training distribution they share, rather than any one model's quirks. We audit where and
why adult-trained affect models fail on child datasets; this design isolates cross-model
child-domain failures rather than a capture-matched causal effect of age.

\begin{figure*}[t]
\centering
\includegraphics[width=0.99\linewidth]{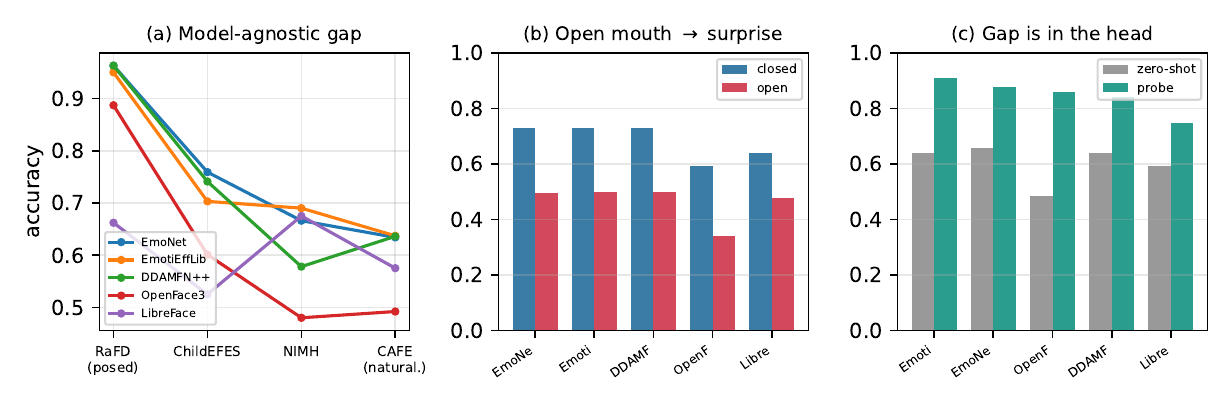}
\caption{The child-affect gap is model-agnostic, demographically concentrated and
AU-grounded, and lives in the classifier head, not the representation.
\textbf{(a)}~Every model degrades from posed (RaFD) to naturalistic (CAFE) faces.
\textbf{(b)}~Open-mouth faces are misread (the AU26 jaw drop $\to$ surprise), all five models.
\textbf{(c)}~A linear probe on frozen features (green) far exceeds zero-shot (grey): the gap is
in the head, not the representation.}
\label{fig:teaser}
\end{figure*}

Our analysis yields a coherent story (\cref{fig:teaser}):

\begin{itemize}
\itemsep0.2em
\item \textbf{The child gap is model-agnostic.} All five architectures degrade from posed to
naturalistic faces (every model far worse on CAFE than RaFD), share the fear$\rightarrow$surprise
confusion, and drop to barely-above-chance signal on spontaneous video (\cref{sec:benchmark}): a
shared cause in the data, not any one model.

\item \textbf{Degradation is concentrated, corroborated, and \emph{not} skin tone.}
Across all five models accuracy drops on open-mouth faces and South-Asian children (smaller
averted-gaze penalty), but not on closed-mouth faces, White/Black children, or direct gaze; the
contrasts hold under participant-level tests (\cref{sec:bias}). The open-mouth$\rightarrow$surprise
error \emph{correlates} with AU26 (jaw drop; monotone dose--response, seen in a second model's
AUs). Black and White children are read equally well, so the bias is not explained by skin tone;
the lower South-Asian accuracy we attribute, as a hypothesis, to training under-representation.

\item \textbf{The gap is the head, not the features.} A linear probe on frozen features reaches
\probeRange{} on \emph{unseen} children versus \zeroshotRange{} zero-shot (\cref{sec:probe}), so
adult-pretrained representations already encode child expressions and the gap lies primarily in
the adult-calibrated head. Dimensional V/A \emph{regression}, by contrast, degrades sharply under
domain shift while the categorical channel is far more recoverable.

\item \textbf{The diagnosis is actionable.} A frozen-backbone head recalibration on a few dozen
target subjects recovers most of the gap and repairs the specific AU26 mechanism with negligible
adult cost, so the audit pinpoints a remedy and bounds its cost.
\end{itemize}

Our primary contribution is the first shared-harness, five-model \emph{audit and benchmark} of
child facial affect: a demographically sliced, AU-grounded bias analysis corroborated across
architectures, and a head-not-features diagnosis verified by a minimal intervention. We will release the harness, per-sample predictions, and
analysis code; the child data stays under its original licenses and is never redistributed: we
release the \emph{audit}, not the data.

\section{Related Work and Background}
\label{sec:related}

\paragraph{Facial expression models.}
We audit five openly available AffectNet-pretrained systems spanning the common design points
(\cref{tab:training}): \textbf{EmoNet}~\cite{toisoul2021emonet} (alignment encoder + 8-way
expression and continuous V/A), \textbf{EmotiEffLib}/HSEmotion~\cite{savchenko2021hsemotion}
(VGGFace2 EfficientNet, multi-task expression+V/A, ABAW lineage), \textbf{DDAMFN++}~\cite{zhang2023ddamfn}
(MS-Celeb-1M dual-attention net), \textbf{OpenFace~3.0}~\cite{openface3} (multi-task
emotion+gaze+AU, VGGFace2 backbone), and \textbf{LibreFace}~\cite{chang2024libreface} (distilled
ResNet-18 for expression+AUs, AffectNet+RAF-DB).

Crucially for an audit of child data, we found \emph{no documented} training on RaFD/Radboud or
on any curated child-face set: every emotion corpus and backbone in their released training
provenance is adult-dominated and in-the-wild (AffectNet~\cite{mollahosseini2017affectnet},
RAF-DB~\cite{li2017rafdb}, FERPlus~\cite{barsoum2016ferplus}, AFEW~\cite{dhall2012afew}, AffWild2~\cite{kollias2023abaw})
or a face-recognition set (VGGFace2~\cite{cao2018vggface2}, MS-Celeb-1M~\cite{guo2016msceleb},
FFHQ~\cite{karras2019ffhq}). Our child datasets, including the
posed RaFD-Kid subset, are therefore held out as far as released provenance shows, and near-ceiling
posed-child accuracy is better explained by ``posed apex faces are easy'' than by known training overlap.

\begin{table}[t]
\centering
\footnotesize
\setlength{\tabcolsep}{4pt}
\begin{tabular}{@{}llccc@{}}
\toprule
Model & Backbone & Emotion train & RaFD? & Child? \\
\midrule
EmoNet      & FAN enc.   & AffN        & no & no \\
EmotiEffLib & VGGFace2   & AffN (+V/A) & no & no \\
DDAMFN++    & MS-Celeb   & AffN-8      & no & no \\
OpenFace~3.0& VGGFace2   & AffN        & no & no \\
LibreFace   & MAE        & AffN+RAF-DB & no & no \\
\bottomrule
\end{tabular}
\caption{Training-data provenance of the five audited models (weights as used; AffN${=}$AffectNet).
Columns list backbone, emotion training corpus, and whether RaFD or any curated child data
appears in the documented training provenance (none for all).}
\label{tab:training}
\end{table}

\paragraph{Child facial-affect datasets.}
We use four child \emph{image} sets, CAFE~\cite{lobue2015cafe} (1192 images, 7 expressions,
demographic labels for race/gender/age and open-/closed-mouth), ChildEFES~\cite{negrao2021childefes},
NIMH-ChEFS~\cite{egger2011nimhchefs} (with direct/averted gaze), and the child subset of the
Radboud Faces Database (RaFD-Kid)~\cite{langner2010rafd}, and two spontaneous child
\emph{video} sets, LIRIS-CSE~\cite{khan2019liris} and EmoReact~\cite{nojavanasghari2016emoreact}.
We chose these to span the axes along which a child audit can break. \emph{Age}: from
preschoolers (CAFE, 2--8\,yr) through school-age children (NIMH-ChEFS, 10--17\,yr; RaFD-Kid) to
the toddler-to-teen range of the spontaneous videos. \emph{Ethnicity}: CAFE and ChildEFES carry
explicit race labels (White, Black, South-Asian, East-Asian, Latino), so demographic gaps can be
\emph{measured} rather than assumed, essential for separating skin tone from expression
morphology (\cref{sec:bias}). \emph{Capture condition}: posed studio frontal (RaFD), gaze-controlled
portraits (NIMH, direct/averted), naturalistic open-/closed-mouth portraits (CAFE), and induced or
fully spontaneous reactions (LIRIS, EmoReact). \emph{Label quality}: all are validated,
peer-reviewed corpora with reported human agreement (CAFE's expressions are validated by
untrained-adult ratings; NIMH and RaFD by the authors' validation studies), so misclassifications
are more readily attributable to the model than to noisy ground truth, though categorical labels
on elicited affect retain residual ambiguity we do not eliminate. EmoReact is multi-label (per-clip
affect-state presences) and carries continuous valence; LIRIS is single-label with no neutral
class.

\paragraph{Adult affect benchmarks.}
For dimensional affect we use AffectNet~\cite{mollahosseini2017affectnet} (valence/arousal +
8 expressions) and the in-the-wild ABAW/Aff-Wild2 challenge~\cite{kollias2023abaw}, whose
Valence-Arousal track provides dense per-frame valence/arousal in $[-1,1]$ with pre-aligned crops.
We pair them deliberately. AffectNet is the \emph{training} distribution of every model we audit,
so it is the in-domain reference that isolates the child gap from any generic generalization gap.
ABAW supplies what no child set can: dense, expert-annotated, continuous valence/arousal under
in-the-wild conditions, on \emph{adults}. It is therefore our \emph{cross-domain} V/A hold-out: a
clean adult-to-adult domain shift that lets us separate ``valence/arousal regression does not
transfer across domains'' from ``\dots\ does not transfer to children,'' which child data alone
(noisy, clip-level, no dimensional labels) cannot disentangle.

\paragraph{Bias and fairness in FER.}
Demographic bias in face analysis is well documented for verification and attribute
prediction, most influentially the intersectional gender-classification gaps of Gender
Shades~\cite{buolamwini2018gendershades}. For expression recognition specifically, prior work
reports accuracy gaps across gender, race, and skin tone~\cite{xu2020fairnessfer} and audits
emotion recognizers for demographic bias~\cite{domnich2021genderbias}, but typically on adults,
for a single model, and without isolating a mechanism. Age bias specifically has been documented
across adult age groups~\cite{kim2021agebias,gayamorey2025agegap}, and child-FER studies remain
largely single-model and single-dataset~\cite{vivekananthan2024children,virgolin2022adults}.
A separate, infant-focused line of work (action-unit detectors~\cite{hammal:et:al:2017} and
PyAFAR-based tools~\cite{hinduja:et:al:2023,ertugrul:et:al:2024}, whose AUs have been used to infer
infant affect~\cite{trinhammer:et:al:2026}) targets infant action units or coarse affect rather than
auditing the adult-pretrained models actually deployed on children.
Remedies have been proposed (adult-data augmentation~\cite{virgolin2022adults} and age-weighted
losses~\cite{gayamorey2025agegap}), but for single models and without localizing the failure; our
minimal head recalibration instead serves as a \emph{diagnostic} that pins the gap to the classifier
head rather than the representation. We contribute a \emph{multi-model,
mechanism-grounded} account on children: we
show which face attributes degrade \emph{consistently across architectures}, tie the
dominant one to a specific action unit, and demonstrate that the cause is largely a
classifier-head miscalibration rather than a representational deficit.
\Cref{tab:priorwork} situates this against prior work: no earlier study combines multi-model scope,
an AU-grounded mechanism, and a head-not-features diagnosis with a deployable fix.

\begin{table*}[t]
\centering
\footnotesize
\setlength{\tabcolsep}{5pt}
\renewcommand{\arraystretch}{1.05}
\begin{tabular}{@{}llccccc@{}}
\toprule
Study & Domain & \#Models & Subgroup slices & AU mech. & Head/feat. & Fix \\
\midrule
Gender Shades~\cite{buolamwini2018gendershades} & adult (gender classif.) & 3 comm. & skin type, gender & $\times$ & $\times$ & benchmark \\
Xu et al.~\cite{xu2020fairnessfer} & FER & 1 & gender, race, age & $\times$ & $\times$ & disentangle \\
Domnich \& Anbarjafari~\cite{domnich2021genderbias} & FER & 5 & gender & $\times$ & $\times$ & $\times$ \\
Kim et al.~\cite{kim2021agebias} & adult FER & 4 comm. & age & $\times$ & $\times$ & $\times$ \\
Gaya-Morey et al.~\cite{gayamorey2025agegap} & FER (age groups) & 3 & age & $\times$ & $\times$ & age-bias mitig.\ \\
Virgolin et al.~\cite{virgolin2022adults} & child FER & 1 & $\times$ & $\times$ & $\times$ & contrastive aug.\ \\
Vivekananthan~\cite{vivekananthan2024children} & child FER & 1 & $\times$ & $\times$ & $\times$ & synthetic aug.\ \\
\midrule
\textbf{Ours} & \textbf{child FER} & \textbf{5} & \textbf{mouth, race, gaze, age, gender} & \checkmark & \checkmark & \textbf{head recalib.} \\
\bottomrule
\end{tabular}
\caption{This audit versus prior facial-affect bias and child-FER work. No prior study grounds the
error in action units or separates a classifier-\emph{head} failure from a \emph{representation}
failure; we add both to a multi-model, demographically sliced child evaluation with a deployable
fix. \#Models counts distinct architectures or commercial (comm.) systems, not training/setup
variants of one backbone.}
\label{tab:priorwork}
\end{table*}

\section{Experimental Setup}
\label{sec:setup}

To make five heterogeneous models comparable, we wrap each behind a common interface and
route all of them through one evaluation pipeline; only the per-model preprocessing differs,
matching each model's training convention.

\paragraph{Common label space.}
Every dataset's labels are mapped to the canonical AffectNet-8 taxonomy (neutral, happy, sad,
surprise, fear, disgust, anger, contempt). Datasets lacking a class are scored only on the
classes they contain. Each model's native output order is remapped to this space; we assert
the mapping against each package's own class list and verify it with a per-class
soft smoke test, since a silent reordering would corrupt accuracy (e.g.\ EmotiEffLib's native
order is alphabetical, DDAMFN uses ``Angry'').

\paragraph{Per-model preprocessing.}
Models own their crop: EmoNet uses a FAN-landmark crop at 256; EmotiEffLib and DDAMFN++ use a
face-bbox crop; OpenFace~3.0 and LibreFace self-detect (bundled RetinaFace / MediaPipe). To
control for detector variance on high-resolution child images (where OpenFace's bundled
detector returns degenerate crops), we feed the shared FAN crop to the bbox models and to
OpenFace, and let LibreFace self-align. All five run in a single environment with per-model
native preprocessing, summarized in \cref{tab:training}.

\paragraph{Metrics.}
For classification we report accuracy and macro-F1 in the main tables (balanced accuracy tracks
accuracy closely). For video we aggregate frame predictions to a per-clip label (majority vote)
and report clip-level accuracy against a majority-class baseline. For dimensional affect we
report the concordance correlation coefficient (CCC) for valence and arousal, and their mean
(the mean-CCC scored by the ABAW valence/arousal challenge). Demographic slices (mouth open/closed, race, gender, gaze, age) are
computed per model from the per-sample predictions; we test contrasts with a two-proportion
$z$-test per model, a participant-level cluster bootstrap (resampling children, not images, to
avoid pseudoreplication), and a sign test across models for corroboration.

\paragraph{Protocol note (AffectNet).}
EmoNet's frequently cited 0.75 AffectNet-8 accuracy is computed on a \emph{cleaned} 2733-image
subset with flip-test-time-augmentation; it is not comparable to the modern models' numbers on
the raw 4000-image validation set. We therefore evaluate \emph{all} models on the raw,
uncleaned, single-crop AffectNet-8 validation set for the fair comparison, and keep EmoNet's
cleaned reproduction only as a weight-validation gate. Four models are scored on the 3999
successfully decoded images; LibreFace is scored on the 3974 it successfully
detected, so its accuracy is recognition-conditional.

\section{Benchmark}
\label{sec:benchmark}

\begin{table}[t]
\centering
\footnotesize
\setlength{\tabcolsep}{3pt}
\begin{tabular}{@{}lccccc@{}}
\toprule
& RaFD & ChildEFES & NIMH & CAFE & mean \\
Model & (80) & (158) & (533) & (1192) & \\
\midrule
EmoNet      & \textbf{.963}\,(.96) & \textbf{.759}\,(.73) & .666\,(.46) & .634\,(.54) & \textbf{.756}\,(.68) \\
EmotiEffLib & .950\,(.95) & .703\,(.64) & \textbf{.690}\,(.47) & \textbf{.637}\,(.63) & .745\,(.67) \\
DDAMFN++    & \textbf{.963}\,(.96) & .741\,(.70) & .578\,(.40) & .636\,(.55) & .730\,(.66) \\
OpenFace~3.0& .887\,(.88) & .601\,(.58) & .480\,(.36) & .492\,(.42) & .615\,(.56) \\
LibreFace   & .662\,(.57) & .525\,(.39) & .675\,(.56) & .575\,(.50) & .609\,(.51) \\
\bottomrule
\end{tabular}
\caption{Child \textbf{image} datasets: accuracy, with \emph{macro-F1} in parentheses (8-class,
mapped; bold marks the best accuracy per column). ``mean'' is the \emph{unweighted} mean over the
four datasets (which have very different $N$; under sample-weighting the top two means swap). Type,
left to right: posed studio frontal $\rightarrow$ posed/induced $\rightarrow$ posed $\pm$gaze
$\rightarrow$ varied open/closed mouth. Every model degrades with naturalness; macro-F1 falls
further on the harder sets (NIMH, CAFE), so accuracy \emph{understates} the gap. No model escapes it.}
\label{tab:childimg}
\end{table}

\begin{table}[t]
\centering
\small
\setlength{\tabcolsep}{4pt}
\begin{tabular}{@{}lcccc@{}}
\toprule
Model & Acc-8 & Acc-7 & Val.\,CCC & Aro.\,CCC \\
\midrule
EmotiEffLib & \textbf{.675} & .675 & .581 & .585 \\
DDAMFN++    & .643 & .654 & -- & -- \\
EmoNet      & .594 & .591 & \textbf{.684} & \textbf{.628} \\
OpenFace~3.0& .562 & .536 & -- & -- \\
LibreFace   & .436 & .498 & -- & -- \\
\bottomrule
\end{tabular}
\caption{Standard AffectNet-8 \emph{validation} (raw 4000, single-crop, no TTA), the fair
cross-model bar. Acc-8/Acc-7 are 8- and 7-class accuracy; Val./Aro.\,CCC are valence/arousal
concordance (dash where a model has no V/A head).}
\label{tab:an8}
\end{table}

\paragraph{Images.}
\cref{tab:childimg} reports accuracy on the four child image datasets. Two patterns hold for
\emph{all five} models. First, accuracy falls from posed to naturalistic faces: from posed
studio RaFD (0.66--0.96, four of five near-ceiling) down to varied open/closed-mouth CAFE
(0.49--0.64); every model scores far below its RaFD accuracy on CAFE, though the drop is not
strictly monotone across the intermediate sets. Second, one per-class failure recurs across
\emph{all} architectures (fear read as surprise), while others, such as anger$\to$disgust, vary
widely by model ($0.00$ for LibreFace to $0.63$ for OpenFace) and are not a shared signature.
EmoNet has the highest \emph{unweighted}-mean accuracy (boosted by the posed sets; the ranking
flips under sample-weighting, where EmotiEffLib leads), and EmotiEffLib is the most robust on the
two hardest sets (NIMH, CAFE); DDAMFN++ ties EmoNet on clean
posed faces but is the most brittle to gaze/naturalness; OpenFace and the distilled LibreFace
trail. The spread across models on any single dataset is typically smaller than the spread across
datasets; i.e., the dataset (domain) explains much of the variance.

\paragraph{Adult reference.}
On the standard AffectNet-8 validation set (\cref{tab:an8}) the modern models cluster at
0.44--0.68, with our reproductions matching published numbers (DDAMFN++ .643 vs .647;
LibreFace Acc-7 .498 vs .497), validating the harness. EmoNet's raw-validation accuracy (.594)
falls squarely in this band; its widely cited 0.75 is an artifact of a cleaned subset plus
flip augmentation, not a real margin over the others.

\paragraph{Video.}
Spontaneous child video is far harder, for reasons intrinsic to the data: the expression is
elicited (unknown onset, most frames neutral/transitional), its intensity uncontrolled, children
move and talk (so the largest deformations come from speech, not the held expression), and the
label is clip-level while the model sees single frames. On LIRIS-CSE (6-class, clip-level majority
vote) the majority baseline (always ``happy'') is 0.342; the five models score 0.30--0.42 (best
fixed window per model), EmotiEffLib's best run exceeding it by only $\sim$0.08, and a data-driven
apex-frame selector did \emph{not} help (maximal deformation coincides with speech/head motion,
not the held expression). On EmoReact (per-clip multi-label presences, threshold-free per-state
ROC-AUC) macro AUC is 0.54--0.62, only just above chance, with happiness the only state detected
appreciably above it. We conclude that frame-aggregated
image models are near chance on spontaneous child video; this needs temporal models, not smarter
single-frame
selection.

\section{Bias Analysis}
\label{sec:bias}

\begin{table}[t]
\centering
\footnotesize
\setlength{\tabcolsep}{4pt}
\begin{tabular}{@{}llccc@{}}
\toprule
Attribute & Hard subgroup & control & \#mdl$\downarrow$ & sig. \\
\midrule
Mouth (CAFE) & open \textbf{.46} & .69 & \textbf{5/5} & $p{<}.001$ \\
Race (CAFE)  & S.\,Asian \textbf{.53} & W.61/B.60 & \textbf{5/5} & $p{=}.02$ \\
Gaze (NIMH)  & averted & direct & \textbf{5/5} & dir. \\
\bottomrule
\end{tabular}
\caption{Demographic slices, pooled over the five models. Columns: hard subgroup accuracy versus
its control, ``\#mdl$\downarrow$'' (models scoring below their own overall accuracy on the
subgroup), and significance.}
\label{tab:bias}
\end{table}

\begin{figure}[t]
\centering
\includegraphics[width=0.82\linewidth]{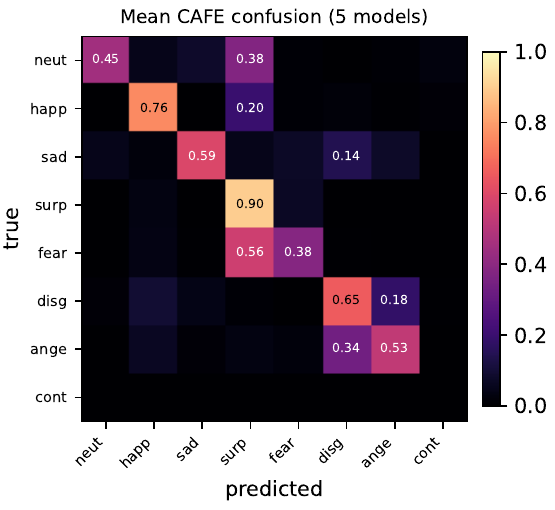}
\caption{CAFE confusion averaged (row-normalized) over all five models. The
fear$\rightarrow$\emph{surprise} confusion is shared across models (5-model mean $0.56$ vs $0.38$
correct); other errors such as anger$\rightarrow$\emph{disgust} are heterogeneous across models
($0.00$--$0.63$) and shown here only as the 5-model average ($0.34$). (CAFE has no contempt.)}
\label{fig:confusion}
\end{figure}

Aggregating per-sample predictions over demographic slices (\cref{tab:bias}) shows the
degradation is not diffuse: it concentrates on \emph{specific} face attributes, and the same
attributes hurt \emph{every} model. Because CAFE and NIMH contain several images per child, we
test each contrast both per image (two-proportion $z$-test) \emph{and} at the participant level
(cluster bootstrap over children), reporting the latter to avoid pseudoreplication. Open-mouth
faces drop to 0.46 (vs 0.69 closed) for all five models (each $p<10^{-3}$ per model; $p=0.0002$
in the participant bootstrap over 146 children): by far the dominant effect. Because surprise is
posed only with a closed mouth in CAFE, this raw contrast mixes class distributions; the
class-standardized gap (macro-averaged over the six classes present in both subsets) is
$+0.10$ to $+0.21$ across models, so the penalty is not an artifact of class mix. South-Asian
children are the lowest-scoring race group (0.53 vs White 0.61, Black 0.60; $n{=}10$ children,
participant-clustered $p=0.02$, exploratory). Averted gaze carries a \emph{small} penalty (direct exceeds
averted by $0.06$ over 52 children, $p\approx0.011$), in the same direction for all five models.
Closed-mouth faces, White/Black children, direct gaze, gender, and most ages are \emph{not}
degraded.

The bias is \emph{not} explained by skin tone. Black and White children, near the extremes of
the skin-tone range, are read equally well (0.60 vs 0.61, difference $-0.003$, $p=0.88$ by
child), so a melanin/reflectance account, which would penalize darker faces, does not fit. (We
use CAFE's race labels, not measured skin tone.) The lower-scoring group is South-Asian. We
\emph{hypothesize} these errors reflect coverage gaps in the adult pretraining data: (i)~the
open-mouth jaw drop (below), a morphology the adult-trained head rarely saw except with surprise,
and (ii)~under-representation of South-Asian faces in in-the-wild adult corpora, rather than a
pixel-level skin-tone effect. Consistent with this, a head recalibrated on a sample of the target
population removes the errors (\cref{sec:method}).

\paragraph{The dominant failure is a jaw drop read as surprise.}
We localize the open-mouth bias mechanistically. Pooled over the five models, 65\% of all
open-mouth misclassifications are predicted as \emph{surprise}, and the surprise
over-prediction rate jumps from 0.06--0.14 (closed) to 0.22--0.51 (open) for every model.
Using LibreFace's per-sample action-unit intensities on CAFE, the per-sample
``fraction of models predicting surprise'' \emph{correlates} with \textbf{AU26 (jaw drop)}~\cite{ekman1978facs}
intensity (Spearman $\rho=0.50$, tie-corrected), rising monotonically across intensity terciles,
0.07 (low) $\rightarrow$ 0.21 (mid) $\rightarrow$ 0.51 (high); AU25 (lips part) is weaker
($\rho=0.21$). The same association appears in a \emph{second} model: OpenFace~3.0 exposes eight
labelled AUs, and its \textbf{AU26 (jaw drop)} intensity correlates with the surprise misfire
($\rho=0.49$), closely matching LibreFace's AU26 ($\rho=0.50$), while its AU25 (lips part) is again
weaker ($\rho=0.26$). Both rest on model-\emph{predicted} AUs (not ground-truth annotations) and
agree by label, so the association holds across two models. The South-Asian error, by contrast, is diffuse (surprise 43\%, disgust 20\%, anger
15\%), indicating a different, distributional cause rather than a single confusion.

\begin{figure}[t]
\centering
\includegraphics[width=\linewidth]{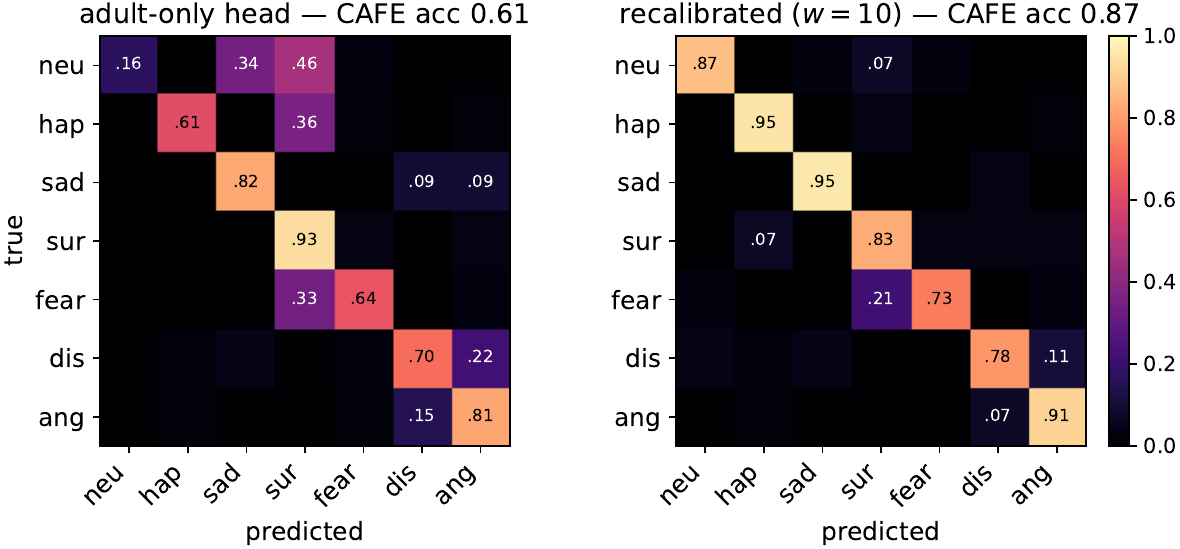}
\caption{Bias and fix are one head miscalibration (EmotiEffLib, held-out CAFE; cf.\
\cref{tab:fix}). The adult-only head over-predicts \emph{surprise}, leaking neutral, happy, and
fear faces into it (left, the AU26 jaw-drop boundary); the child-recalibrated head ($w{=}10$)
collapses that mass onto the diagonal (CAFE $0.61\to0.87$, backbone frozen; right).}
\label{fig:recalconf}
\end{figure}

\section{Probe Analysis}
\label{sec:probe}

\begin{table}[t]
\centering
\small
\setlength{\tabcolsep}{6pt}
\begin{tabular}{@{}lccc@{}}
\toprule
Model & zero-shot & linear probe & $\Delta$ \\
\midrule
EmotiEffLib & .638 & \textbf{.910} & $+.272$ \\
EmoNet      & .657 & \textbf{.878} & $+.221$ \\
OpenFace~3.0& .484 & \textbf{.859} & $+.375$ \\
DDAMFN++    & .638 & \textbf{.843} & $+.205$ \\
LibreFace   & .593 & \textbf{.747} & $+.154$ \\
\bottomrule
\end{tabular}
\caption{Linear probe on \emph{frozen} features, CAFE, \textbf{participant-disjoint}
(108 train / 46 test subjects), all five models. Columns: zero-shot accuracy, accuracy of a
linear head probe on the frozen features, and their difference $\Delta$.}
\label{tab:probe}
\end{table}

If adult-pretrained \emph{representations} were the problem, no cheap fix would exist. They
are not. We freeze each backbone, extract embeddings, and train only a linear
classifier~\cite{alain2016probes} on a \emph{participant-disjoint} CAFE split (no child appears in both train and test), then compare
to zero-shot accuracy on the same held-out children (\cref{tab:probe}). Across all five models a
linear head reaches 0.75--0.91 on unseen children, versus 0.48--0.66 zero-shot ($\Delta = +0.15$
to $+0.38$). The clearest evidence is OpenFace~3.0: it has the worst zero-shot accuracy (0.484) yet
its features probe to 0.859, as well as any other model. The child-FER gap is therefore largely a
\emph{classifier-head} miscalibration to adult statistics rather than a representational deficit:
the features already encode child expressions well enough for a linear head to recover
0.75--0.91, above the 66\% mean accuracy untrained adults reach on CAFE~\cite{lobue2015cafe}: the
signal is recoverable, so the representation is not the bottleneck; the adult-trained boundary
simply does not read it. (The probe identifies the head as the \emph{dominant}, not sole, factor;
\cref{tab:fix} corroborates across four child sets.) This both explains the
open-mouth/AU26 confusion (an adult-calibrated boundary places the jaw-drop region near
``surprise'') and points to a cheap remedy: refit the head, do not fine-tune the backbone.

\paragraph{Dimensional affect tells the opposite, sharper story.}
Valence/arousal \emph{regression} degrades sharply under domain shift. EmoNet's valence CCC
is 0.68 on the AffectNet validation set (in-domain), but only 0.21 on the adult in-the-wild
ABAW hold-out and 0.22 on child EmoReact (degraded but nonzero). The adult cross-domain drop is
\emph{as large as} the child drop, so poor child valence is a property of dimensional regression
under domain shift, not something child-specific. Categorical expression labels, in contrast, are
far more \emph{recoverable} from frozen features after target adaptation (the probe above). This
dissociation, that \emph{categorical structure is more recoverable than dimensional regression
under the same shift}, motivates routing affect through
the categorical channel, which we develop next.

\section{Method: Recalibrate the Head}
\label{sec:method}

The probe (\cref{sec:probe}) says the child expression gap is a head miscalibration, so a head
refit on a little child data should recover it cheaply, with the backbone frozen, preserving
adult performance and the auxiliary heads (valence/arousal, gaze, AUs).

\paragraph{Pooled head recalibration.}
We retrain only the linear expression head on frozen features, pooled over a class-balanced
sample of AffectNet-train (adults) and the child datasets, and evaluate on held-out adults
(AffectNet-val) \emph{and} held-out children (subject-disjoint), sweeping the child up-weight
to trace the adult/child frontier. The fix is architecture-agnostic: it works on every one of
the five models (\cref{tab:fix}). At $w{=}10$, CAFE improves by $+0.21$ to $+0.27$ and NIMH by
$+0.13$ to $+0.28$ on all five backbones, while the \emph{adult cost is negligible} (at most $-0.015$
versus the adult-only refit, far smaller than the child gains): the child gains are essentially
free. The two small posed sets are noisier: ChildEFES gains on three of five models, and RaFD
is already near ceiling. Because the backbone is frozen, the valence/arousal, gaze, and
action-unit heads are untouched. This is a deployable fix: collect a little child expression
data, refit one linear layer. Crucially, it repairs the \emph{specific} mechanism \cref{sec:bias}
localized, not just the average: on a held-out CAFE split the recalibrated EmotiEffLib head lifts
open-mouth accuracy from $0.49$ to $0.85$ and cuts the open-mouth$\to$surprise misfire from $0.38$
to $0.07$, while fear accuracy rises $0.49\to0.70$ as the fear$\to$surprise confusion more than
halves (all relative to the adult-only refit head): the bias and the fix are the same head
miscalibration (\cref{fig:recalconf}).

\begin{table}[t]
\centering
\footnotesize
\setlength{\tabcolsep}{3.5pt}
\begin{tabular}{@{}llccccc@{}}
\toprule
model & head & adults & CAFE & ChildEFES & NIMH & RaFD \\
\midrule
\multirow{2}{*}{EmotiEffLib} & adult-only & .683 & .606 & .667 & .704 & .958 \\
 & $w{=}10$ & .678 & \textbf{.865} & \textbf{.789} & \textbf{.858} & .958 \\
\midrule
\multirow{2}{*}{EmoNet}      & adult-only & .610 & .657 & .737 & .747 & .958 \\
 & $w{=}10$ & .595 & \textbf{.865} & .702 & \textbf{.914} & .917 \\
\midrule
\multirow{2}{*}{DDAMFN++}    & adult-only & .639 & .590 & .649 & .599 & .750 \\
 & $w{=}10$ & .632 & \textbf{.859} & \textbf{.754} & \textbf{.784} & \textbf{1.00} \\
\midrule
\multirow{2}{*}{OpenFace~3}  & adult-only & .584 & .603 & .614 & .556 & .958 \\
 & $w{=}10$ & .575 & \textbf{.849} & .596 & \textbf{.833} & .958 \\
\midrule
\multirow{2}{*}{LibreFace}   & adult-only & .570 & .580 & .579 & .648 & .792 \\
 & $w{=}10$ & .560 & \textbf{.811} & \textbf{.632} & \textbf{.778} & .792 \\
\bottomrule
\end{tabular}
\caption{Frozen-backbone head recalibration across all five models: a linear expression head
retrained on pooled AffectNet-train + child data (child up-weight $w$), evaluated on held-out
adults (AffectNet-val) and held-out, subject-disjoint children. Each model has two rows: the
adult-only baseline (head retrained without child data) and the recalibrated head at $w{=}10$.
Bold marks the recalibrated ($w{=}10$) result where it improves on the adult-only baseline.}
\label{tab:fix}
\end{table}

\paragraph{The gain is in-distribution.} The fix recalibrates to the \emph{target} distribution,
not a generic ``child'' head. Training on three child datasets and testing on the held-out fourth
transfers weakly: mean improvement over the adult-only head is $+0.05$/$+0.04$ for
DDAMFN++/OpenFace~3, $\approx 0$ for EmotiEffLib, and \emph{negative} for EmoNet and LibreFace
(pooling other child sets even costs LibreFace $-0.18$ on NIMH), far short of the $+0.2$
in-distribution gain. In frozen feature space the four child datasets shift each expression's
class centroid in \emph{different} directions (a class-conditional domain
shift~\cite{wang2018dasurvey}, not a feature deficit), so a head fitted to one collection lands on
the wrong boundary for another. The recommendation sharpens: collect a little expression data
\emph{from the deployment population and capture conditions}.

\begin{figure}[t]
\centering
\includegraphics[width=0.86\linewidth]{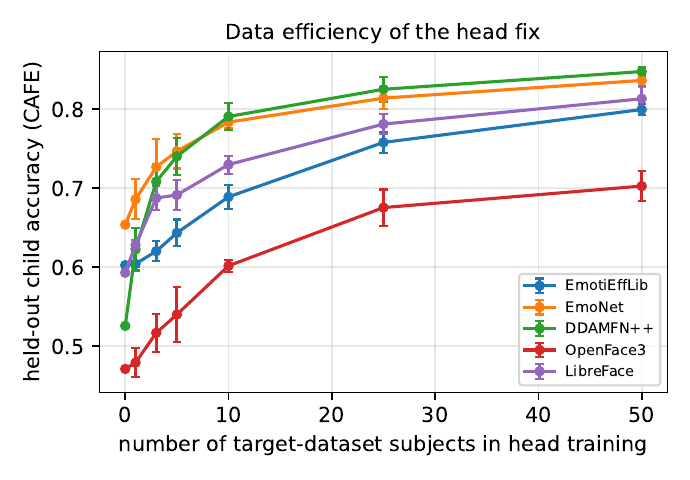}
\caption{Data efficiency of the head recalibration on CAFE: held-out child accuracy versus the
number of target subjects used to refit the linear head (8k-frame AffectNet anchor, child
up-weight $w{=}10$; mean$\pm$std over 4 subject draws). Most of the gain arrives by 10--25
children.}
\label{fig:dataeff}
\end{figure}

\paragraph{How little target data?} Very little (\cref{fig:dataeff}): sweeping target subjects on
CAFE, accuracy climbs steeply, with most of the gain captured by $\sim$10--25 children and
diminishing (still positive) returns out to the 50 we test; even 5 subjects lift most models above
their zero-shot floor. The unit is \emph{children}, not images: each contributes about 8 CAFE
frames, so $\sim$25 children is $\sim$190 images, class-balanced and up-weighted ($w{=}10$, i.e.\
$\sim$1.9k in effective weight) so they are not swamped by the 8k-frame adult anchor. This is less
surprising than it sounds: we refit only a \emph{linear} head, and the probe (\cref{sec:probe})
already showed the frozen features encode child expressions, so the head needs only a small
boundary correction. A few dozen labelled children from the target setting suffice.

\begin{figure}[t]
\centering
\begin{subfigure}{\linewidth}
\centering
\includegraphics[width=0.82\linewidth]{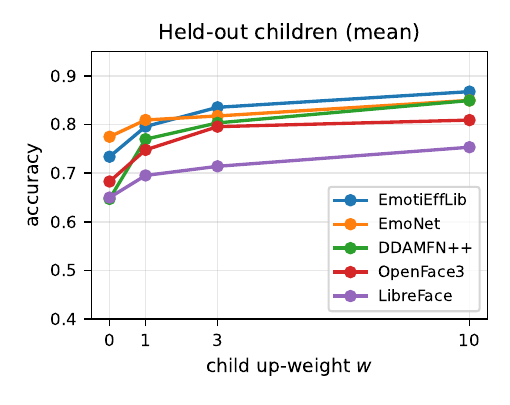}
\caption{Held-out children climb steeply.}\label{fig:frontier-child}
\end{subfigure}
\\[2pt]
\begin{subfigure}{\linewidth}
\centering
\includegraphics[width=0.82\linewidth]{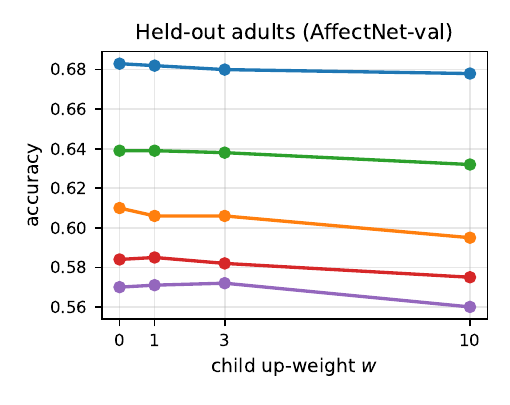}
\caption{Held-out adults stay flat.}\label{fig:frontier-adult}
\end{subfigure}
\caption{The head recalibration buys child accuracy essentially for free. As the child up-weight
$w$ rises, held-out \emph{children} improve sharply on all five backbones \subref{fig:frontier-child}
while held-out \emph{adults} stay nearly flat \subref{fig:frontier-adult} (at most $-0.015$): the
adult cost is negligible relative to the child gains.}
\label{fig:frontier}
\end{figure}

\paragraph{Valence/arousal is harder, and reveals a dissociation.}
Available child corpora lack dense per-frame V/A labels comparable to the adult benchmarks, so
child V/A can only be \emph{transferred}. On EmotiEffLib (the strongest native V/A model, scored by
valence-CCC on the adult ABAW hold-out) we compare three post-hoc routes against its native,
jointly optimized head (0.375): regress from frozen features (0.218), map the categorical posterior
(probs$\to$V/A, 0.250), or fine-tune the backbone (0.276). The posterior route beats feature
regression, the \emph{same} categorical-over-dimensional advantage the probe showed, but \emph{no}
route reaches the native head (scaling the map to 138k adult frames plateaus at 0.335; fusing the
degraded predicted AUs adds nothing). The dissociation holds, yet closing the gap to the native
cross-domain V/A head is the open problem: more adult data does not close it, and the binding
constraint is the absence of comparable dimensional \emph{child} labels.

\paragraph{Why not fine-tune the backbone?}
Fine-tuning the last block$+$head (1.1M params, with AffectNet rehearsal and a masked adult-only
V/A loss) loses to the frozen head on the only two models with native V/A heads (EmotiEffLib and
EmoNet; \cref{tab:an8}):
child accuracy rises less (CAFE 0.80 vs the frozen 0.865) while adult V/A drifts below native
($\sim$0.28 vs 0.375), and refitting V/A on the fine-tuned features does not recover it. We test
these two because the key cost (V/A-head degradation) is observable only on V/A-capable models;
the frozen-head recalibration already covers all five (\cref{tab:fix}). Across the tested
configurations, fine-tuning did not beat the frozen head, consistent
with the probe's verdict that the gap lives in the head, not the representation.

\section{Discussion}
\label{sec:discussion}

\paragraph{Limitations and future work.}
Two boundaries scope the result and point to the next steps. The head fix is in-distribution by
construction: it recalibrates to the target population and does not transfer across child
collections, so \emph{predicting} transferability from a few capture-condition descriptors would
turn it into a broader recipe that says in advance when one population's data can stand in for
another's. And child valence/arousal lacks dimensional labels comparable to the adult benchmarks,
so carrying the categorical-channel dissociation onto the dimensional axis remains the hardest open
problem the audit surfaces. Beyond these, frame-aggregated models set a clear baseline on spontaneous child
video, motivating temporal architectures that model expression dynamics; and the head-not-features
framework is not specific to children, so the same audit and recalibration could test other
under-represented populations (elderly faces, clinical groups, cross-cultural settings).

\paragraph{Ethics and societal impact.}
The child datasets were collected and released by their creators under dataset-specific consent
and ethics procedures; we use them in place under their licenses, contact no participants, add no
new data collection, and never upload or redistribute images, releasing only code, per-sample
outputs, and aggregate statistics. Affect predictions are not ground-truth internal states and
should not drive diagnostic, disciplinary, or access decisions: any deployment requires consent,
human oversight, and validation in the target population. We used LLM coding assistants (Claude,
Codex) for code and writing; no labels, data, or experimental results were LLM-generated, and all
decisions and responsibility rest with the authors.

\section{Conclusion}
\label{sec:conclusion}

A controlled audit of five adult-trained models on child data shows the gap is model-agnostic,
demographically concentrated and corroborated across architectures (open mouths read as surprise,
correlating with the AU26 jaw drop; South-Asian children lower; but not explained by skin tone),
and largely a classifier-head miscalibration rather than a representational deficit. The probe and
recalibration experiments identify adult-label calibration as the dominant failure: large-scale
pretraining already exposes the backbone to diverse faces, so it encodes child expressions, while
a little child-labeled data repairs the decision boundary. Training-set underrepresentation is a
plausible source of that miscalibration. The practical consequence is a
cheap, deployable remedy, refitting one linear layer on a few dozen labelled children, but an
\emph{in-distribution} one: the gains do not transfer across child collections, so the data must
come from the deployment population and capture conditions. Dimensional valence/arousal is harder
still and remains substantially less transferable, the open problem we surface.

\section*{Acknowledgments}
This work was partially funded by the Deutsche For\-schungs\-ge\-mein\-schaft (DFG, German Research Foundation)~-- project number~490909448.\par

{
    \small
    \bibliographystyle{ieeenat_fullname}
    \bibliography{main}
}
\end{document}